\pdfoutput=1

\documentclass[11pt]{article}

\usepackage[final]{acl}

\usepackage{times}
\usepackage{latexsym}

\usepackage[T1]{fontenc}

\usepackage[utf8]{inputenc}

\usepackage{microtype}

\usepackage{inconsolata}

\usepackage{graphicx}

\usepackage{tabularx}
\usepackage{booktabs}
\usepackage{amsmath}
\usepackage{array}
\usepackage{stfloats}

\usepackage[ruled,vlined]{algorithm2e}
\usepackage{algpseudocode}

\title{LLM Optimization Unlocks Real-Time Pairwise Reranking}

\author{
  Jingyu Wu, Aditya Shrivastava, Jing Zhu, Alfy Samuel, Anoop Kumar, Daben Liu \\
  AI Foundations, Capital One \\
  \{jingyu.wu, aditya.shrivastava2, jing.zhu, alfy.samuel, anoop.kumar, daben.liu\}@capitalone.com
}

\begin{document}
\maketitle

\begin{abstract}
Efficiently reranking documents retrieved from information retrieval (IR) pipelines to enhance overall quality of Retrieval-Augmented Generation (RAG) system remains an important yet challenging problem. Recent studies have highlighted the importance of Large Language Models (LLMs) in reranking tasks. In particular, Pairwise Reranking Prompting (PRP) has emerged as a promising plug-and-play approach due to its usability and effectiveness. However, the inherent complexity of the algorithm, coupled with the high computational demands and latency incurred due to LLMs, raises concerns about its feasibility in real-time applications. To address these challenges, this paper presents a focused study on pairwise reranking, demonstrating that carefully applied optimization methods can significantly mitigate these issues. By implementing these methods, we achieve a remarkable latency reduction of up to 166 times, from 61.36 seconds to 0.37 seconds per query, with an insignificant drop in performance measured by Recall@k. Our study highlights the importance of design choices that were previously overlooked, such as using smaller models, limiting the reranked set, using lower precision, reducing positional bias with one-directional order inference, and restricting output tokens. These optimizations make LLM-based reranking substantially more efficient and feasible for latency-sensitive, real-world deployments.

\end{abstract}

\section{Introduction}

Efficient and accurate retrieval systems are critical for delivering relevant content in industrial applications such as customer support~\cite{10.1145/3626772.3661370}, finance, and banking~\cite{zhao2024revolutionizing}. Retrieval-Augmented Generation (RAG) pipelines, which integrate retrieval and generation can enhance response quality. Large Language Models (LLMs) have demonstrated significant potential in optimizing these pipelines through reranking ~\cite{zhu2023large}. However, their high computational costs and latency present substantial barriers to widespread deployment.

Pairwise Reranking Prompting (PRP)~\cite{qin-etal-2024-large}, which ranks candidate documents through pairwise comparisons, effectively enhances relevance. However, its extensive computational requirements make it impractical for many real-time applications. This necessitates optimization strategies that maintain accuracy while reducing computational overhead.

In this paper, we present an optimized system (Figure \ref{fig:system_design}) with PRP for industrial-scale retrieval-augmented generation applications, aiming to lower latency without compromising reranking performance. We  evaluate our approaches on two real-world proprietary datasets. Our optimizations include leveraging smaller LLMs like FLAN-T5-XL, implementing single pass of sliding window for the most salient document reranking, constraining reranking scope to a carefully chosen \emph{Top-K}, loading LLMs with lower precision, applying one-directional order inference to mitigate positional bias, and using constrained single-token generation. Collectively, these methods reduce inference time from over 60 seconds to 0.37 seconds per query using four A100 GPUs, with minimal impact on model performance.

Our main contributions are: \begin{itemize} \item Demonstrating that carefully optimized LLM-based reranking is viable for real-time industrial applications. \item Providing a set of generalizable latency-reduction techniques for LLM-based tasks that enables more effective, real-time retrieval systems in diverse domains where relevant information must be delivered promptly and accurately. \end{itemize}

\section{Background}
\subsection{LLM-Based Text Reranking}
LLMs have driven significant advances in text reranking \cite{zhu2023large}. Traditional approaches, predominantly listwise (e.g., RankGPT \cite{sun2023chatgptgoodsearchinvestigating}, LRL \cite{ma2023zero}) and pointwise (e.g., \cite{liang2022holistic}, \cite{sachan2022improving}), have inherent drawbacks. Listwise methods, which rank all documents simultaneously, typically underperform unless powered by highly capable LLMs such as GPT-4. By contrast, pointwise methods, which assess each document's relevance independently, struggle to generate well-calibrated scores without fine-tuning. Recent exploration of setwise approaches~\cite{setwise2024} has shown promise for efficiency, but still lags behind pairwise methods in performance. Given these challenges, particularly when performance cannot be traded for speed, Pairwise Reranking Prompting (PRP)~\cite{qin-etal-2024-large} emerges as a compelling solution. PRP leverages LLMs' comparative strengths to rerank documents effectively without additional training, making it a practical choice for real-world applications, especially if latency improvements can be realized.

\subsection{Advancements in Pairwise Reranking Prompting}
Pairwise Reranking Prompting (PRP), introduced by \cite{qin-etal-2024-large}, has rapidly gained traction in the field of information retrieval because of the off-the-shelf usability of the pretrained LLMs without additional finetuning. Its applications have extended beyond ranking to encompass broader evaluation tasks \cite{liusie-etal-2024-llm, park2024paireval}. While Instruction Distillation \cite{sun2023instruction} has shown promise, its effectiveness appears limited to LLMs of GPT-4 scale or larger. Efforts to refine PRP have yielded approaches like LLM-RankFusion \cite{zeng2024llm}, which addresses inconsistencies in PRP by incorporating additional context. However, this increased context length introduces latency concerns. Similarly, PRPGraph \cite{luo-etal-2024-prp} enhances PRP's performance but also faces latency bottlenecks.  Despite widespread acknowledgment of latency limitations in pairwise methods \cite{liu2024aligning, pradeep2023rankzephyr, liu2024leveraging, zhao2023survey, zhu2023large},  mitigating these challenges without compromising performance remains an open research area.




\begin{figure*}[!ht]
  \includegraphics[width=\textwidth]{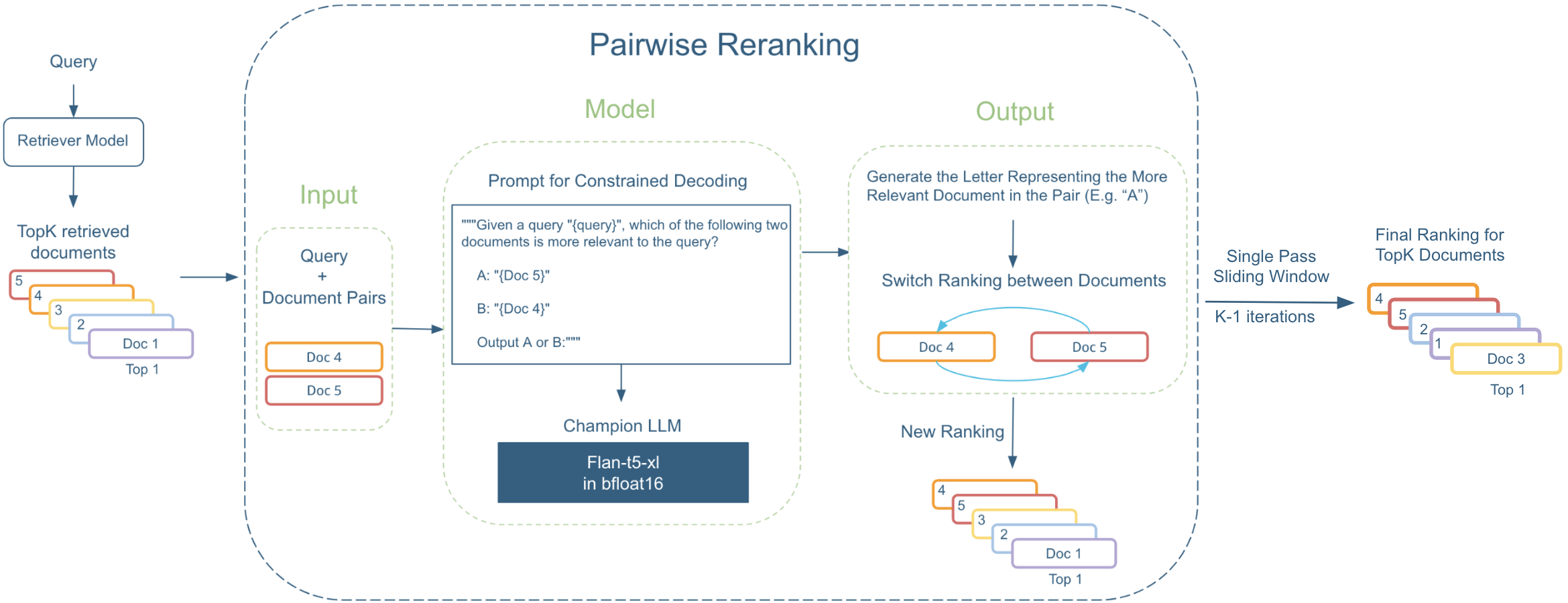}
  \caption{The Optimized Pairwise Reranking System}
  \label{fig:system_design}
\end{figure*}

\section{Method} \label{methods}
In this section, we provide an in-depth description of the optimization techniques explored to improve the inference latency of the PRP methodology. 

\subsection{Model Size Reduction}
We first analyze the importance of model size in the Pairwise Reranking approach. Our experiments primarily use the FLAN-T5 \cite{raffel2020exploring} family of models (FLAN-T5-XXL (11.3B), FLAN-T5-XL (2.85B), and FLAN-T5-Large (783M)). We additionally extend the comparison to FLAN-UL2 (20B) \cite{tay2022ul2}, which is also a FLAN-T5 based model. The core step in Pairwise Reranking involves comparing the relevance of two documents with respect to a given user query. 

Based on our findings, we conclude that for relatively simple tasks—such as text comparisons requiring concise outputs—larger models do not necessarily offer additional performance benefits compared to smaller models. In contrast, using a smaller model can substantially improve efficiency by reducing latency and resource requirements. 

\subsection{Sliding Window with Single Pass}
Traditional reranking methods resemble full sorting, where the retrieved documents are reranked based on their relevance with the query. However, such global reordering incurs substantial computational overhead and proves redundant when only a concentrated set of highly relevant results is sought.

Our approach employs a sliding window mechanism \cite{qin-etal-2024-large} with a single pass to iteratively identifying the most salient document, significantly reduces computational complexity to O(n). 
While extended to identify a small subset of highly ranked documents, our experimental results (Table \ref{tab:topk}) consistently indicate that optimizing beyond the single highest-ranked document (top-1) yields only marginal improvements in recall rate. This suggests the initial highest-ranked document captures predominant relevant information, making top-1 optimization the most resource-efficient strategy for our datasets.

\vspace{-3pt}
\begin{table}[ht]
\centering
\setlength{\tabcolsep}{2pt}
\begin{tabular}{lcccccccc}
\hline
\text{Rerank} & \multicolumn{3}{c}{\text{Recall@k} \footnotesize(Dataset 1)}  & & \multicolumn{3}{c}{\text{Recall@k} \footnotesize(Dataset 2)} & \text{Latency}\\ 
\text{Goal} & \text{k=5} & \text{k=3} & \text{k=1} && \text{k=5} & \text{k=3} & \text{k=1} &  \text{(s)} \\ \hline
Top-5 & 0.77 & 0.73 & 0.56 && 0.79 & 0.75 & 0.54 & 1.97 \\ 
Top-3 & 0.76 & 0.73 & 0.56 && 0.79 & 0.75 & 0.54 & 1.06 \\ 
Top-1 & 0.75 & 0.71 & 0.56 && 0.77 & 0.71 & 0.54 & \textbf{0.37} \\ \hline
\end{tabular}
\caption{Sliding Window on Different Reranking Goals}
\label{tab:topk}
\end{table}


\subsection{Optimizing the Reranking Threshold \textit{Top-K}}
The reranking stage, irrespective of the underlying algorithm, introduces a hyperparameter, \textit{Top-K}, which specifies how many of the retriever’s top-ranked results are subsequently passed to the reranker. In our investigation, we systematically test \textit{Top-K} to measure its impact on both retrieval accuracy and efficiency. We find that while increasing \textit{Top-K} can improve accuracy up to a certain extent, any further increase yields diminishing returns. 

In practice, an appropriate \textit{Top-K} value can be identified using grid search, binary search, or simple trial-and-error. Once recall rate meets the desired performance level, the additional computational cost of reranking more candidates typically outweighs the marginal performance gains. Hence, pinpointing the ``sweet spot'' that optimizes both efficiency and accuracy becomes crucial.

\subsection{Loading Models with Lower Precision}
We further explore lowering the floating-point precision from \texttt{float32} to \texttt{float16} or \texttt{bfloat16} when loading model weights, reducing computational cost without substantially degrading performance.

Our findings with different precision settings highlight that even modest reductions from \texttt{float32} to \texttt{bfloat16} can notably decrease inference time.

Separately, we also explore quantization, which maps model weights from floating-point to lower-precision integer representations (e.g., 8-bit or 4-bit). While quantization can provide additional speedups beyond mere precision reduction, we observe that it introduces more substantial changes to the numerical representations, resulting in a noticeable performance decline. Given these trade-offs, we opt not to include quantization in our final implementation.

\subsection{One-directional Order Inference}

Recent studies have shown that LLMs are prone to positional bias, exhibiting sensitivity to the order in which input elements are presented \cite{qin-etal-2024-large, zeng2024llm, zhao-etal-2024-measuring, luo-etal-2024-prp}. A common mitigation strategy involves performing pairwise comparisons in both directions to neutralize input-order effects.

In this work, we investigate a latency-efficient alternative using LLMs by adopting a one-directional order inference scheme with deliberate position assignment. Specifically, during inference, we consistently designate the document with the lower initial retrieval ranking as input A and the higher-ranked document as input B. This configuration ensures that the model processes the lower-ranked document first, which we hypothesize reduces the tendency to favor the first input, thereby partially mitigating positional bias while improving efficiency.

\subsection{Constrained Decoding}
In pairwise reranking, most of the inference time is spent generating output tokens. To mitigate this, we constrain the model to produce exactly one token (e.g., ``A'' or ``B'') through prompt engineering, thereby reducing inference overhead. However, simply taking the first token generated as the final output can lead to invalid or irrelevant response unless the prompt is carefully designed.

To address this, we systematically test multiple prompt variations and select the one (in Figure \ref{fig:system_design}) that yields the highest accuracy under single-token constraints. 
Algorithm~\ref{alg:constrained-decoding} outlines our approach: we evaluate each candidate prompt on a test set, record its accuracy, and select the best-performing prompt for constrained decoding. Combined with greedy decoding at zero temperature, this approach achieves performance 
on par with multi-token outputs while substantially reducing computational time.

\begin{algorithm}[ht]
\SetAlgoLined
\KwIn{Model $M$, Test Set $D=\{(c^A_i, c^B_i, y_i)\}$, Prompt Candidates $P$}
\KwOut{Best Prompt $p^*$ with Single-Token Decoding}

Initialize $\text{bestAccuracy} \leftarrow 0$, $p^* \leftarrow \text{None}$\;

\ForEach{$p \in P$}{
    correctCount $\leftarrow 0$\;
    \ForEach{$(c^A, c^B, y)$ in $D$}{
        fullPrompt $\leftarrow \text{formatPrompt}(p, c^A, c^B)$\;
        outputToken $\leftarrow \text{generate}(M, \text{prompt} = fullPrompt, \text{max\_new\_tokens}=1, \text{greedy}=\text{True}, \text{temperature}=0)$\;
        \If{outputToken = $y$}{
            correctCount $\leftarrow \text{correctCount} + 1$\;
        }
    }
    accuracy $\leftarrow \frac{\text{correctCount}}{|D|}$\;
    \If{accuracy > \text{bestAccuracy}}{
        $\text{bestAccuracy} \leftarrow \text{accuracy}$, $p^* \leftarrow p$\;
    }
}
\Return{$p^*$}
\caption{Iterative Prompt Selection for Constrained Decoding}
\label{alg:constrained-decoding}
\end{algorithm}

\section{Experiments}

\subsection{Data}
We employ two distinct datasets, dataset 1 and dataset 2, both derived from proprietary internal use cases for our experiments. Each dataset corresponds to a unique business vertical. Within each dataset, the structure comprises:

\begin{itemize} 
\item A document corpus that serves as the primary contextual knowledge base.

\item A test set consisting of user-generated queries, each meticulously paired with its corresponding relevant context from the corpus.
\end{itemize}

\subsubsection{Corpus}
The corpus is the knowledge base for a business vertical. The corpus in dataset 1 consists of 8379 documents, while corpus in dataset 2 consists of 8121 documents. Considering the token limit of the model input, the maximum token length of each document is 512. These documents represent segmented, structured entries derived from real business knowledge bases used in production environments.

\subsubsection{Test Set}
Two test sets are included in the datasets corresponding to the corpora described above. Each test set contains real user queries derived from actual usage of the internal search system. For each query, a relevant context—a segment from the knowledge base that directly answers the query—is provided, along with a corresponding link that identifies the document in the knowledge system where the context is found. Both the context and the link are manually annotated by domain experts for each respective business vertical, serving as ground truth in our experiments. The volume of test sets is 975 queries for dataset 1 and 736 queries for dataset 2.

\begin{table}[h]
\centering
\label{tab:dataset_stats}
\setlength{\tabcolsep}{3pt}
\begin{tabular}{lcc}
\hline
                            & \text{Dataset 1} & \text{Dataset 2} \\ \hline
Total Documents in Corpus   & 8379               & 8121               \\ 
Tokens per Document (Min)   & 192                 & 214                 \\
Tokens per Document (Avg.)  & 474                & 509              \\
Tokens per Document (Max)   & 512               & 512               \\
Total Queries               & 975                & 736                \\ \hline
\end{tabular}
\caption{Statistics on the Two Datasets Used in Our Experiments}
\end{table}

\subsection{Evaluation Metrics}
In our experiments, Recall rate is selected as the primary evaluation metric, aligning directly with our industry-specific use cases. This metric quantifies the proportion of relevant documents successfully retrieved by the system out of the total available documents in the corpus. We specifically utilize Recall@k, as our datasets feature a single correct document as ground truth for each query. This metric directly measures the percentage of queries for which the unique relevant document is successfully identified within the top-k retrieved results. 

Alternative metrics, such as Normalized Discounted Cumulative Gain (NDCG), which are designed for scenarios with multiple relevant documents and graded relevance, are deemed less appropriate and potentially overly complex for our specific single-answer retrieval task.

\subsection{System Settings}
We experiment with different computational resources. The latency on A100 is significantly lower than A10G GPUs. All the experiments reported in this paper are using four A100 GPUs.

\subsection{Retriever Model Setup}
Our baseline retriever model is \textit{multi-qa-mpnet-base-cos-v1}, a sentence-transformer model trained on 215M \{question, answer\} pairs. It is the champion retriever model on our datasets.



\begin{table*}[t]
\centering
\begin{tabular}{lccccccccccc}
\hline
\textbf{Reranking} & \multicolumn{5}{c}{\textbf{Recall@k} (Dataset 1)}  & & \multicolumn{5}{c}{\textbf{Recall@k} (Dataset 2)}\\ 
\textbf{Strategy} & \textbf{k=25} & \textbf{k=10} & \textbf{k=5} & \textbf{k=3} & \textbf{k=1} &  & \textbf{k=25} & \textbf{k=10} & \textbf{k=5} & \textbf{k=3} & \textbf{k=1}\\ \hline
Cross-encoder & 0.90 & 0.85 & 0.80 & 0.73 & 0.52 &  &  0.90 & 0.83 & 0.76 & 0.67 & 0.47\\ 
Pairwise & 0.90 & 0.85 & 0.80 & 0.75 & \textbf{0.58} &  &  0.90 & 0.85 & 0.79 & 0.74 & \textbf{0.54}\\ 
\hline
\end{tabular}
\caption{Pairwise Reranking Outperforms Cross-encoder Reranking on Both Datasets}
\label{tab:compare}
\end{table*}

\subsection{Comparison with Classical Ranking Techniques}
To substantiate the efficacy and necessity of employing pairwise reranking within our system, we conduct a comparative analysis against a more established reranking paradigm. Specifically, we benchmark its performance against cross-encoder reranking, a widely adopted strategy in information retrieval tasks.

Our evaluation involves comparing the performance between the cross-encoder and the pairwise reranker when applies atop the baseline retriever previously described. We use \textit{bge-reranker-v2-m3} as the cross-encoder reranker. At the time of this study's inception, this model is recognized for its state-of-the-art performance in the field of cross-encoder reranking. Its demonstrated capability in achieving high retrieval effectiveness makes it a compelling and robust choice for our comparative analysis. As evidenced in Table~\ref{tab:compare}, the pairwise reranker consistently demonstrates superior performance across both datasets. This significant outperformance underscores the potency of the pairwise reranker for complex reranking tasks.

Despite its inherently higher latency compared to classical cross-encoder reranking, the enhanced effectiveness of the pairwise approach validates its integration into the retrieval pipeline, making it a crucial component for achieving optimal ranking quality.

\section{Results and Discussion}
\subsection{Baseline}
As illustrated in the first half of Table \ref{tab:all}, the baseline application of pairwise reranking produces a marked improvement in Recall@k. For instance, in Dataset 1, Recall@1 jumps from 0.42 (Retriever Only) to 0.58 (Retriever + Pairwise Reranker), confirming that the llm-based comparison of competing candidate documents substantially enhances ranking performance. We observe a similar trend in Dataset 2, where pairwise reranking boosts Recall@1 from 0.50 to 0.54. These improvements underscore the significance of the problem being addressed—enabling more accurate ranking of relevant information despite the inherent challenges posed by large and complex databases.

\begin{table*}[t]
\centering
\setlength{\tabcolsep}{3pt}
\begin{tabular}{lccccccccccc}
\hline
& \multicolumn{4}{c}{\textbf{Recall@k} (Dataset 1)}  & & \multicolumn{4}{c}{\textbf{Recall@k} (Dataset 2)} & \textbf{Latency}  & \textbf{Speedup} \\ 
\textbf{Setup} & \textbf{k=10} & \textbf{k=5} & \textbf{k=3} & \textbf{k=1} &  & \textbf{k=10} & \textbf{k=5} & \textbf{k=3} & \textbf{k=1} &  \textbf{(s)} & \textbf{Factor}\\ \hline
Baseline (Retriever Only) & 0.77 & 0.68 & 0.62 & 0.42 &  &  0.80 & 0.73 & 0.67 & 0.50 & - & - \\ 
+ Pairwise Reranker & 0.85 & 0.80 & 0.75 & \textbf{0.58} &  &  0.85 & 0.79 & 0.74 & \textbf{0.54} & 61.36 & - \\ 
\midrule
+ Flan-UL2 $\rightarrow$ Flan-T5-XL & 0.85 & 0.81 & 0.76 & 0.59 &  & 0.85 & 0.81 & 0.75 & 0.56 & 22.50 & $\sim2.7\times$ \\ 
+ TopK=25 $\rightarrow$ TopK=5 & 0.81 & 0.72 & 0.66 & 0.52 &  & 0.84 & 0.77 & 0.71 & 0.54 & 3.27 & $\sim6.9\times$ \\ 
+ \texttt{float32} $\rightarrow$ \texttt{bfloat16} & 0.81 & 0.72 & 0.66 & 0.52 &  & 0.84 & 0.77 & 0.71 & 0.54 & 2.23 & $\sim1.5\times$ \\ 
+ One-directional Order & 0.81 & 0.72 & 0.66 & 0.52 &  & 0.84 & 0.77 & 0.71 & 0.54 & 1.11 & $\sim2\times$ \\ 
+ Constrained Decoding & 0.81 & 0.72 & 0.66 & 0.52 &  & 0.84 & 0.77 & 0.71 & 0.54 & \textbf{0.37} & $\sim3\times$\\
(Speedup Gain in Total) & &  & & &  & & & & &  & $\sim\textbf{167}\times$ \\\hline
\end{tabular}
\caption{Model Performance, Optimization Methods and Their Corresponding Latency Improvements on Two Datasets}
\label{tab:all}
\end{table*}

\subsection{Optimization Results}
The lower half of Table \ref{tab:all} provides a comprehensive breakdown of the latency improvements achieved through the optimization techniques discussed in Section \ref{methods}. Since the latency values for Dataset 1 and Dataset 2 are approximately equivalent across all methods, we report latency results for Dataset 2 for clarity and conciseness. Each method contributes uniquely to enhancing efficiency of our Pairwise Reranking algorithm. Starting with the replacement of FLAN-UL2 with FLAN-T5-XL, a model that is approximately seven times smaller, we observe 2.7 times improvement in latency. Notably, this change also results in a slight increase in retrieval performance, demonstrating that smaller models can effectively handle tasks requiring concise outputs, such as pairwise comparisons, without compromising accuracy.

Among all the optimization strategies, \textit{Top-K} tuning yields the most significant performance gains. While reducing \textit{Top-K} from 25 to 5 results in a drop in Recall@1 (0.59 to 0.52 on Dataset 1 and 0.56 to 0.54 for Dataset 2), the broader retrieval quality—particularly the consistency of Recall@10—remains stable. This indicates that while fewer candidates are processed at the reranking stage, the model retains its ability to identify the most relevant documents within the top ranked documents. This optimization achieves a 6.9 times improvement in latency. The computational cost saving far outweighs the marginal reduction in top-1 accuracy. We consider this the second most effective optimization, as it balances efficiency and effectiveness for tasks that prioritize recall performance on a small number of top search results.

One of the most meaningful observations is the role of constrained decoding. By enforcing single-token outputs in addition to all other optimization strategies, we lower latency by a factor of 3 times without any recall drop and finally achieve a runtime latency of lower than 1 second–at just 0.37 seconds per query. This improvement fundamentally changes the feasibility of deploying pairwise reranking in real-time applications, where sub-second responsiveness is often critical. Remarkably, this drastic gain in efficiency does not compromise retrieval accuracy, as Recall@1 remains consistent with other optimized configurations. This demonstrates that with careful prompt design and decoding constraints, tasks requiring categorical outputs can achieve near-optimal performance with significantly reduced computational overhead. 

Combined with one-directional order inference, which halves the latency associated with positional bias, and reduces model precision (\texttt{bfloat16}), these strategies collectively transforms the system’s scalability. The cumulative impact of these optimizations underscores a key takeaway: a holistic approach to PRP optimization, targeting both algorithmic bottlenecks and hardware efficiency, is essential and feasible for real-world scalability of LLM-based pairwise reranking solutions.

\section{Conclusion and Future Work}

We address the challenge of integrating Pairwise Reranking Prompting (PRP) into real-world retrieval and RAG systems, showing that careful optimizations can reduce per-query latency from over 60 seconds to well under a second with negligible performance loss. These optimizations—selecting smaller models, restricting the reranked set, adopting lower precision, mitigating positional bias via one-directional order inference, and constraining output tokens—collectively transform PRP into a viable option for latency-sensitive applications.

In future research, we plan to extend our optimization framework to other reranking paradigms—listwise, pointwise, and hybrid—to assess whether similar latency–accuracy trade-offs can be achieved. We also aim to explore dynamic reranking strategies that adjust the Top-K value based on query complexity or system load, potentially improving both recall and responsiveness in practical settings.

While model parallelism offers limited benefit in our current setup—likely due to the relatively modest model size—we see data or pipeline parallelism as promising areas for further investigation. These strategies may offer greater gains under higher-throughput conditions or when scaling to larger models. Finally, we anticipate further performance gains could be realized through model- or task-specific tuning and more effective batching strategies within the reranking pipeline.

By systematically balancing performance and computational efficiency, our approach outlines a robust, scalable blueprint for real-time LLM-based retrieval solutions. In doing so, we bridge the gap between cutting-edge research and practical deployment constraints across diverse industry settings.




\section*{Limitations}
While our proposed optimizations significantly reduce PRP latency and improve deployability in real-world RAG systems, several limitations remain. 

First, our reranking improvements rely on a fixed Top-K strategy. The current system does not adjust reranking scope based on query complexity or load, which may limit efficiency under highly variable conditions.

Second, our reranking pipeline—though optimized—does not yet incorporate task-specific or domain-adaptive tuning. This may constrain performance in specialized applications (e.g., legal, medical) where domain-specific language and retrieval behavior differ significantly from the training regime.

\bibliography{anthology,custom}

\clearpage
\appendix

\section{Appendices}

\subsection{Methods Tried but Not Successful}

\textbf{Model Parallelism}
\label{sec:appendix}

From the model structure perspective, a commonly used inference optimization method is model parallelism. Model parallelism distributes different layers of models in different GPUs and accelerates inference by doing computations on different GPUs parallelly. 
However, in our experiments, model parallelism doesn’t give a huge improvement on latency. The reason behind could be the fact that Flan-T5-XL is small enough to be loaded in one A100 GPU. Since doing computation on only one GPU at a time can save the communication cost between GPUs, smaller models may benefit more when being loaded into one GPUs rather than multiple GPUs. Thus, we decide to pause on model parallelism experiments.








\textbf{Altering Attention Mechanisms}
\label{sec:appendix}

In this section, we describe our efforts to optimize the latency of Pairwise Reranking by incorporating faster attention mechanisms. Specifically, we explore three different attention mechanisms applied to the T5 family of models: SDPA, Eager and Flash Attention 2 \cite{dao2023flashattention}.

We experiment with various attention mechanisms, some of which do not yield significant improvements. Importantly, we find that these mechanisms improve performance with larger models, such as Flan-T5-XXL. However, when we transition to smaller models through prompt engineering, the gains are minimal. This suggests a trade-off between model size and the benefits of faster attention mechanisms. Larger models tend to benefit more, while smaller models may not exhibit the same level of improvement.

\begin{figure}[htbp!]
  \centering
  \begin{minipage}{0.45\textwidth}
    \centering
    \includegraphics[width=\linewidth]{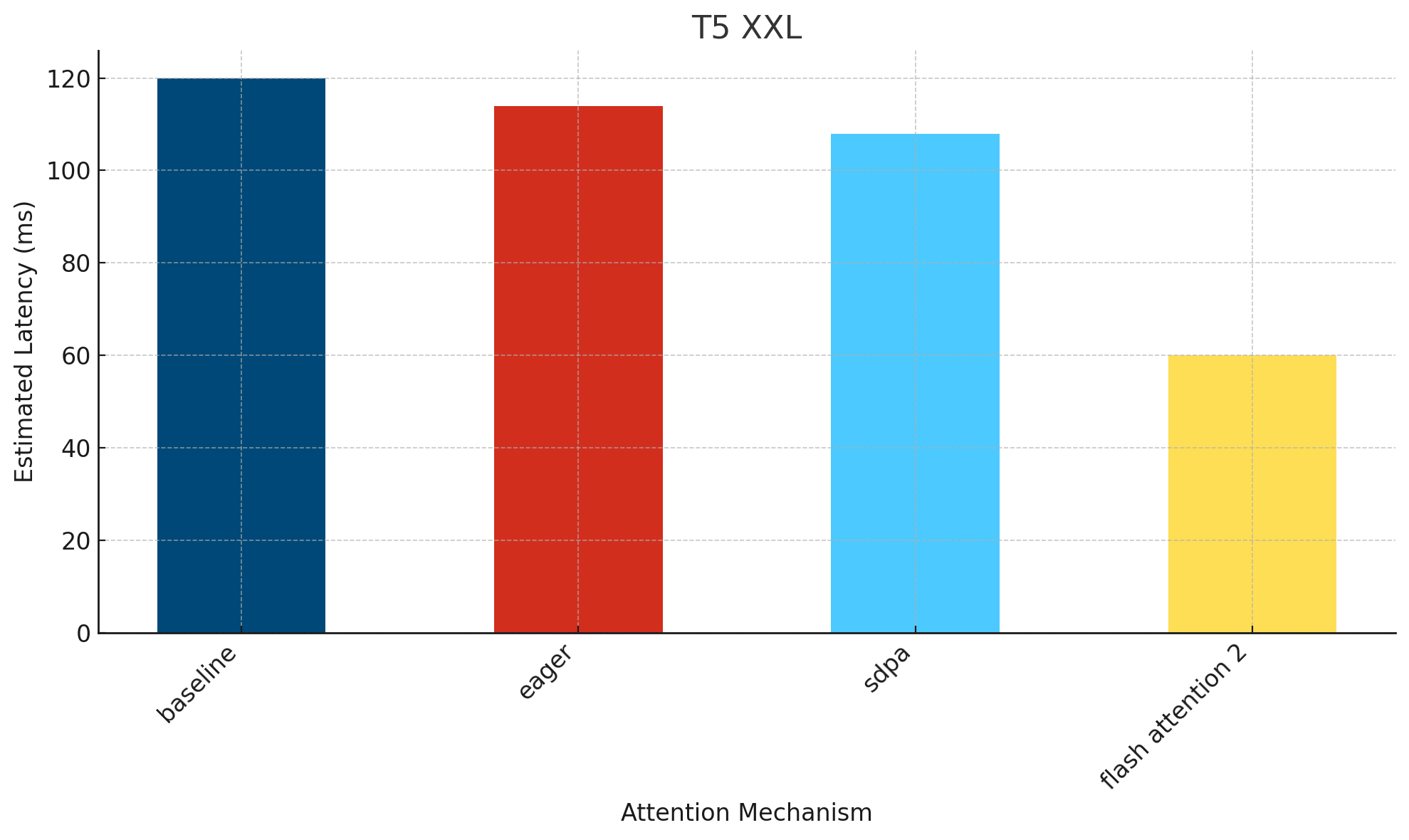}
    \label{fig:t5xxl}
  \end{minipage}\hfill
  \begin{minipage}{0.45\textwidth}
    \centering
    \includegraphics[width=\linewidth]{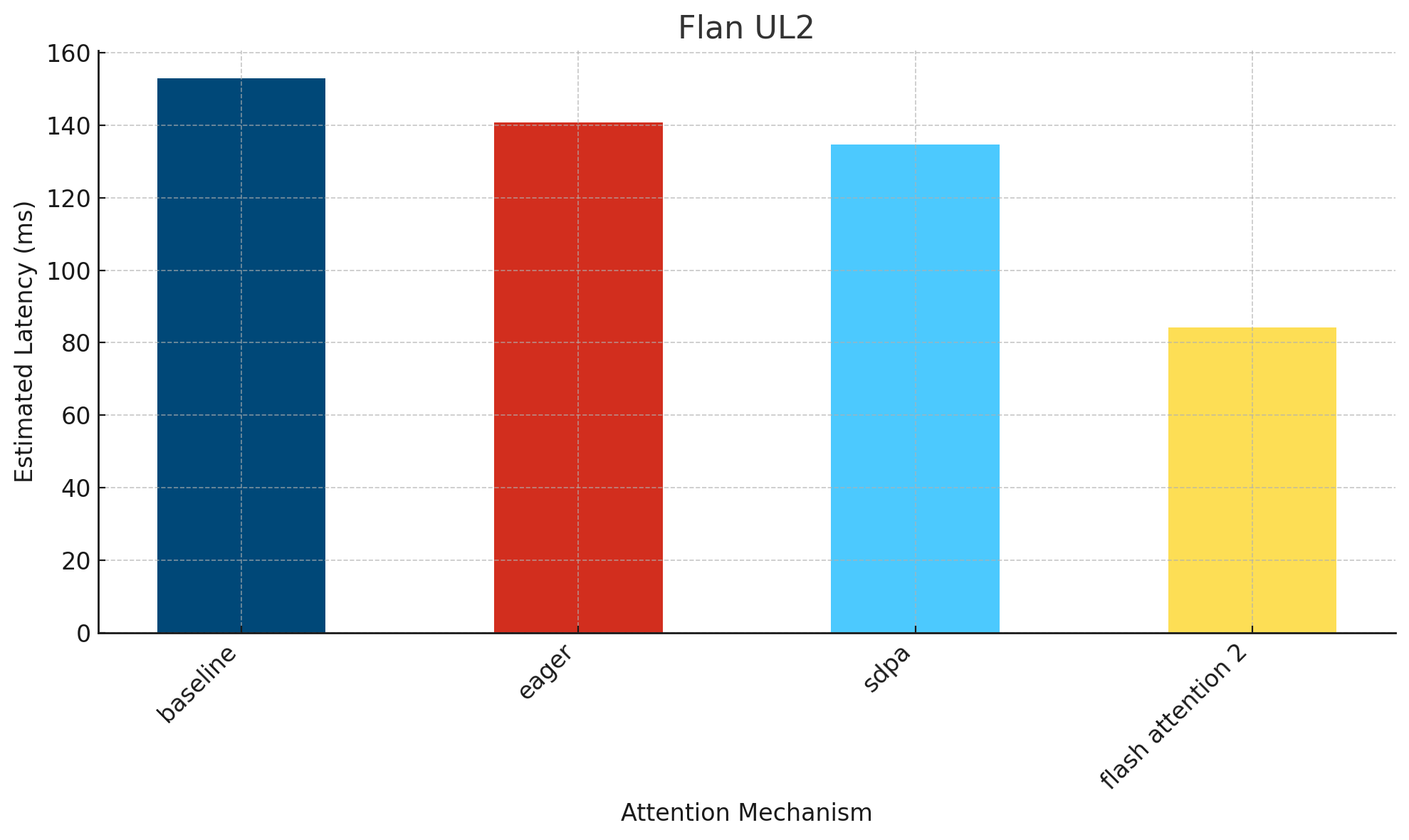}
    \label{fig:flanul2}
  \end{minipage}

  \vspace{0.5cm} 

  \begin{minipage}{0.45\textwidth}
    \centering
    \includegraphics[width=\linewidth]{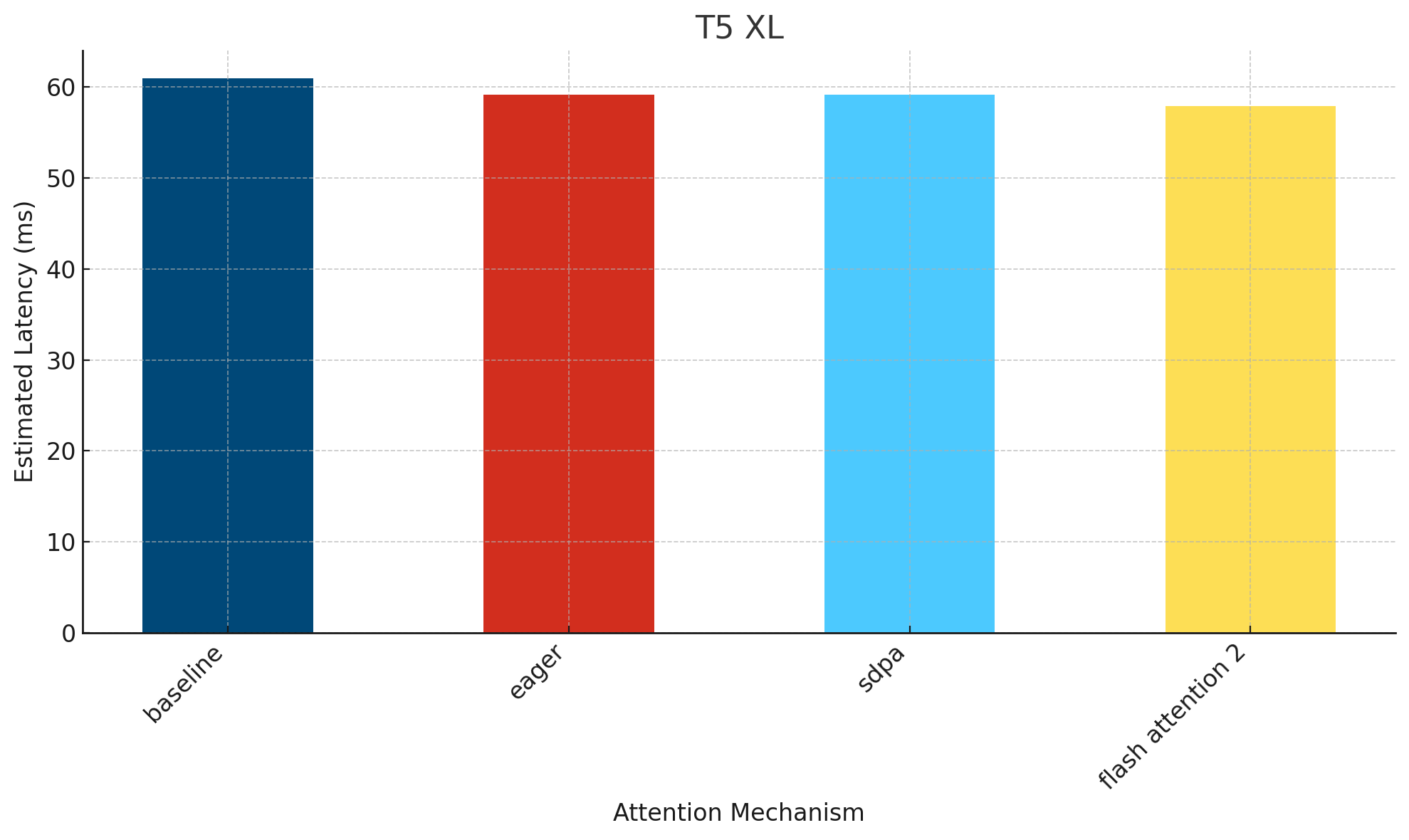}
    \label{fig:flanxl}
  \end{minipage}\hfill
  \begin{minipage}{0.45\textwidth}
    \centering
    \includegraphics[width=\linewidth]{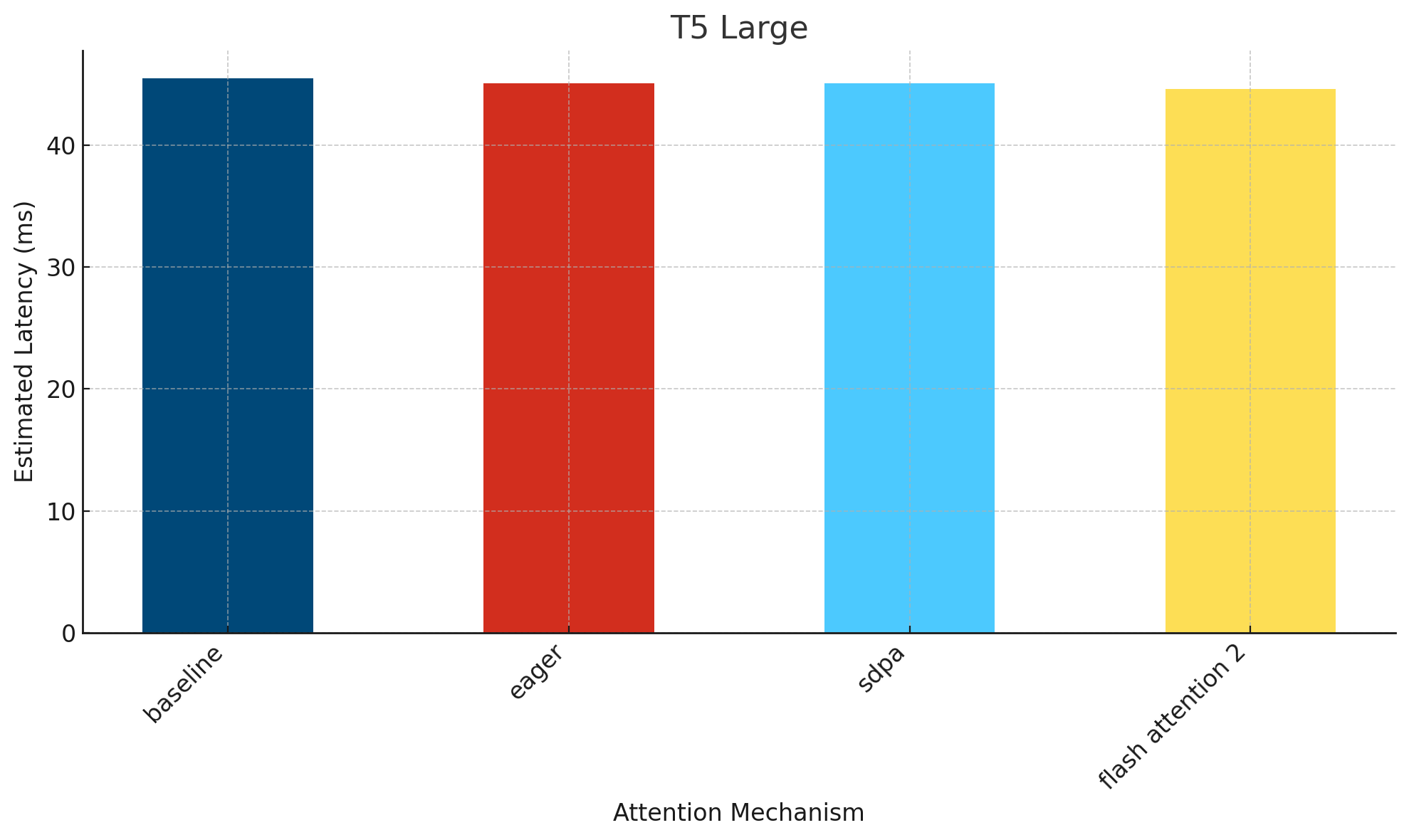}
    \label{fig:smallmodels}
  \end{minipage}

  \caption{Performance comparison in terms of latency per query of different models with faster attention mechanisms (from top to bottom: Flan-T5-XXL, Flan-UL2, Flan-XL, and Flan-T5-Large).}
  \label{fig:grid}
\end{figure}

\textbf{Batch Inference}
\label{sec:appendix}

We also systematically experiment with a range of batch sizes to evaluate the impact on performance. In this procedure, the input batch in the first round contains approximately \( n/2 \) samples or pairs, where \( n \) is the total number of samples in the initial batch. After processing the first batch, the most similar text pairs are promoted to the next round, which reduces the number of samples to \( n/4 \), and so on. This process continues for approximately \( \log_2 n \) rounds, depending on the total number of initial samples.

\subsection{Prompt List}
\label{sec:appendix}

\begin{table*}[!tb]
  \centering
  \begin{tabular}{|p{3cm}|p{12cm}|}
    \hline
    \textbf{Version} & \textbf{Primary Prompt}\\
    \hline
    \text{Original} & """Given a query \{query\}, which of the following two passages is more relevant to the query?

    Passage A: \{doc1\}

    Passage B: \{doc2\}

    Output Passage A or Passage B:"""\\ 
    \hline
    \text{Passage Mode} & """Given a query \{query\}, which of the following two passages is more relevant to the query?

    Passage A: \{doc1\}

    Passage B: \{doc2\}

    Output the more relevant Passage:"""\\ 
    \hline
    \text{Number Mode} & """Given a query \{query\}, which of the following passages is more relevant to the query?

    Passage 1: \{doc1\}
    
    Passage 2: \{doc2\}
    
    The answer is captured in a * format. For example, *2* or *1*. Just return the number of the more relevant Passage:"""\\ 
    \hline
    \text{Final Version} & """Given a query \{query\}, which of the following two passages is more relevant to the query?

    A: \{doc1\}

    B: \{doc2\}

    Output A or B:"""\\ 
    \hline
    \end{tabular}
    \caption{Primary Prompt Variations in Our Experiments}
  \label{tab:prompt}
\end{table*}

Table~\ref{tab:prompt} shows the list of primary variations of prompts that we have experimented with. 

We conduct a systematic evaluation of several prompt variations, ultimately choosing the final version that achieves the highest performance which enables single-token constraints.
\label{sec:appendix}

\end{document}